\documentclass[runningheads]{llncs}
\usepackage[T1]{fontenc}
\usepackage{makecell} 
\usepackage{array}
\usepackage{multirow}
\usepackage{subcaption} 
\usepackage{graphicx}
\usepackage{hyperref}
\usepackage{amsmath}
\begin{document}

% Accelerating Active Inference: 
\title{A Hardware-oriented Approach for Efficient Active Inference Computation and Deployment}
\vspace{-2cm}
% \author{Anonymized for submission}
\author{Nikola Pižurica\inst{2,3} \and
Nikola Milović\inst{3} \and
Igor Jovančević\inst{2,3}  \and
Conor Heins \inst{1}  \and
Miguel de Prado \inst{1} 
}
% \authorrunning{Anonymized for submission}
\authorrunning{N. Pižurica, N. Milović, I. Jovančević, C. Heins, and M. de Prado}

% \institute{}
\institute{VERSES, Los Angeles, California, 90016, USA \and
Computer Science Center, University of Montenegro, 81000 Podgorica, Montenegro \and
%Princeton University, Princeton NJ 08544, USA \and
Fain Tech, 81000 Podgorica, Montenegro \\
\email{nikola.p@ucg.ac.me}
\email{miguel.deprado@verses.ai}}
% \email{\{autho1, author2\}verses.ai}}
\maketitle              % typeset the header of the contribution
\vspace{-0.4cm}
\begin{abstract}
Active Inference (AIF) offers a robust framework for decision-making, %under uncertainty, 
yet its computational and memory demands pose challenges for deployment, especially in resource-constrained environments. This work presents a methodology that facilitates AIF's deployment by integrating \textit{pymdp}'s flexibility and efficiency with a unified, sparse, computational graph tailored for hardware-efficient execution. Our approach reduces latency by over 2x and memory by up to 35\%, advancing the deployment of efficient AIF agents for real-time and embedded applications.
\vspace{-0.3cm}
\end{abstract}

\vspace{-0.8cm}
\section{Introduction}
\vspace{-0.2cm}
Active Inference (AIF) \cite{ActInfBook} is emerging as a powerful paradigm for building intelligent, adaptive agents, grounded in Bayesian inference and variational free energy. Despite its powerful theoretical foundations and growing practical relevance, deploying AIF agents efficiently on hardware (HW) remains challenging, especially in real-time or resource-constrained systems on the \textit{edge}~\cite{serb2017hardware}. 

\textit{Pymdp} \cite{heins2022pymdp} is a flexible Python-based library for prototyping Active Inference agents, offering computational efficiency via its \textit{JAX} backend \cite{jax2018github}. 
However, \textit{Pymdp} suffers from highly unstructured graphs, posing several issues for efficient HW acceleration.
%However, it suffers from limited computational efficiency. 
%In an effort to overcome this limitation, \textit{pymdp} has adopted \textit{JAX} \cite{jax2018github}, a library for high-performance computing, as a backend.  While \textit{JAX} highly improves efficiency through vectorization and just-in-time (jit) compilation, Pymdp's AIF execution graphs remain highly unstructured, posing several issues for efficient HW acceleration.
Other libraries such as \textit{cpp-AIF} (C++) \cite{nehrer2025introducing}, \textit{ActiveInference.jl} \cite{gregoretti2024cpp}, and \textit{RxInfer.jl} \cite{bagaev2023rxinfer} (Julia) have emerged with a strong focus on performance. %though %, enabling faster and more scalable simulations. 
%their adoption has been limited. 
Nonetheless, \textit{cpp-AIF} %, though optimized via multi-core parallelization, 
requires low-level programming expertise and lacks the Python's high-level abstractions, making rapid prototyping and integration more difficult. 
In contrast, \textit{RxInfer.jl} is a powerful, general-purpose Bayesian inference engine but places less particular emphasis on active inference with POMDPs. Moreover, while Julia-based tools are performant and composable, they face adoption challenges due to the language’s relative novelty and smaller user community.
%\textit{RxInfer.jl} is a package for Bayesian inference based on reactive message passing. % in a factor graph representation. 
%\textit{ActiveInference.jl} leverages Julia’s performance and composability with \textit{RxInfer.jl}, offering a novel package for variational Bayesian inference based on reactive message passing. % in a factor graph representation. 
%However, the language’s relative novelty and smaller user base may limit accessibility and community support.
This work proposes to remodel \textit{pymdp} to produce compact,
structured computational graphs, enabling efficient HW mapping and GPU acceleration.

\vspace{-0.3cm}
\section{Methodology}
\vspace{-0.2cm}
\paragraph{\textbf{Principle:}} 
AIF is a principled framework for adaptive behavior grounded in the Free Energy Principle. It enables agents to learn and act under uncertainty by minimizing variational free energy — a proxy for surprise — within a probabilistic generative model \cite{friston2016activelearning}. This model captures the joint distribution over hidden states and observations, allowing %agents 
to infer the latent causes of sensory inputs. %Agents minimize expected free energy, balancing information gain (exploration) and expected utility (exploitation). 

\vspace{-0.3cm}\paragraph{\textbf{Problem formulation:}}
In \texttt{pymdp}, each observation modality $o^m\!\in\!\{1,\dots,L_m\}$ and hidden‐state factor $s^n\!\in\!\{1,\dots,K_n\}$ is linked by a Categorical–Dirichlet pair.  
With $N$ total hidden state factors and $M$ total observation modalities, storing every conditional table \smash{$\{p(o^m\,|\,s^1, s^2, \dots, s^N)\}_{m=1}^M$} requires $ \mathcal{O}\!\bigl(M\,L_{\max}\,K_{\max}^{\,N}\bigr) $
%\[
%\mathcal{O}\!\bigl(M\,L_{\max}\,K_{\max}^{\,N}\bigr)
%\]
memory and an equal order of floating-point operations for each computation that operates on these tensors. 
$L_{\max}$ refers to the maximum cardinality (or alphabet size) across observation modalities, and $K_{\max}$ to the 
%maximum 
cardinality across state factors.
With more than a handful of factors, this “fully enumerated’’ representation becomes prohibitive.

% \medskip
\noindent\textbf{Current implementation:}  
Recent patches\footnote{\href{https://github.com/infer-actively/pymdp/pull/127}{pymdp \#127}} let users specify the pattern of conditional independencies in the generative model.  
Exploiting this \textit{structural sparsity} (the absence of links) avoids allocating tensors that are not relevant for inference, but leaves two problems untouched:

\begin{enumerate}\setlength{\itemsep}{0pt}
\vspace{-0.2cm}
    \item \textbf{Functional sparsity}: even inside the remaining tensors most parameter values are still zero or negligible.
    \item \textbf{Unwieldy computational graphs}: each modality and factor lives in a separate Python list, forcing irregular-shaped nested \texttt{for}-loops during inference and policy rollouts that lead to notable overheads and poor GPU mapping.
\end{enumerate}

\vspace{-0.2cm}

\begin{figure}[t!]
    \centering
    \begin{subfigure}[t]{0.75\textwidth}
        \centering
        \includegraphics[width=\linewidth, height=4.1cm]{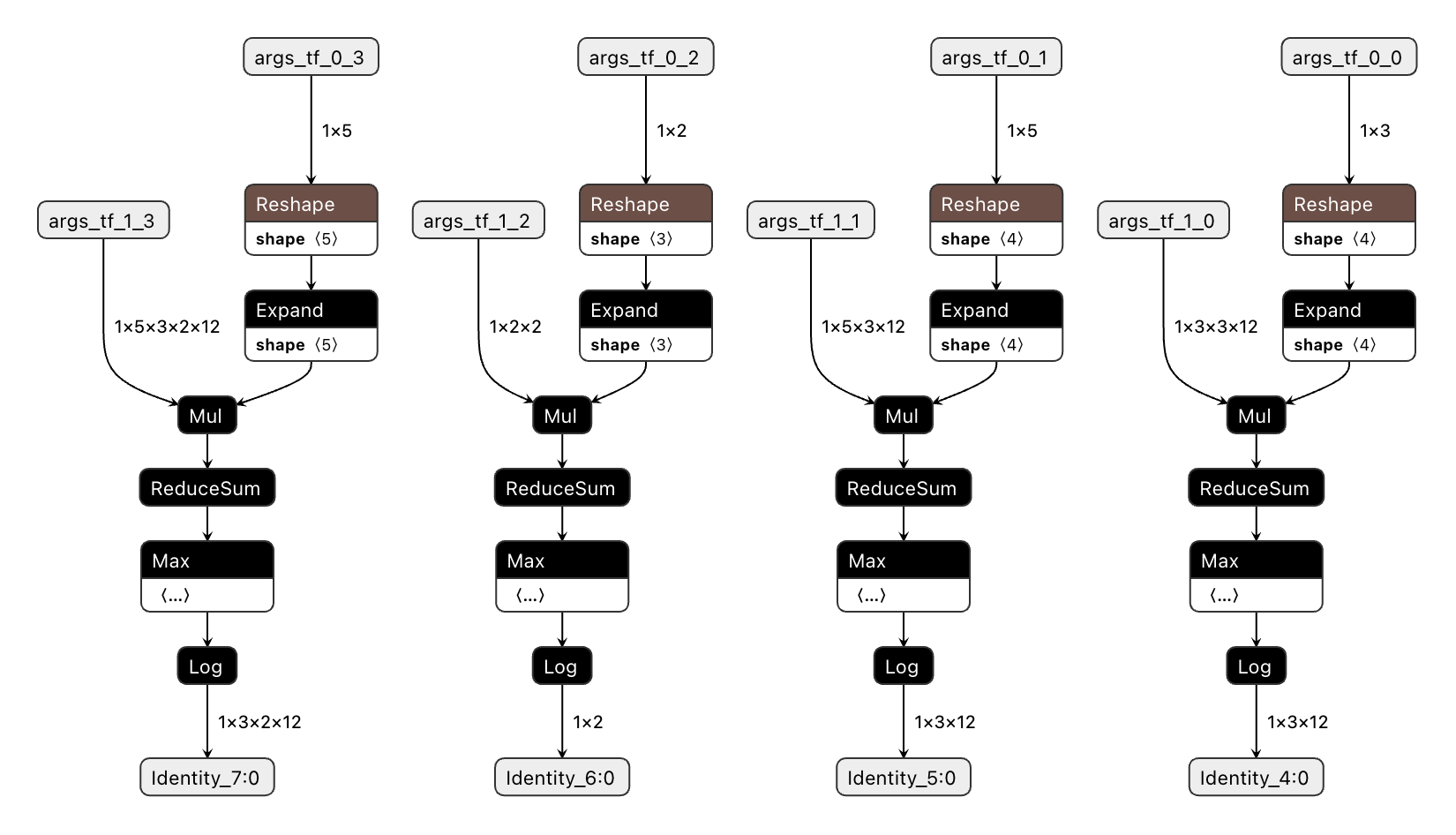}
        \caption{Pymdp's original computational graph (ONNX format) with functional sparsity, forcing irregular-shaped nested \textit{for}-loops.}
        \label{fig:left}
    \end{subfigure}
    \hfill
    \rule{0.5pt}{4cm}  % Thin vertical line
    \begin{subfigure}[t]{0.22\textwidth}
        \centering
        \includegraphics[width=0.8\linewidth, height=4cm]{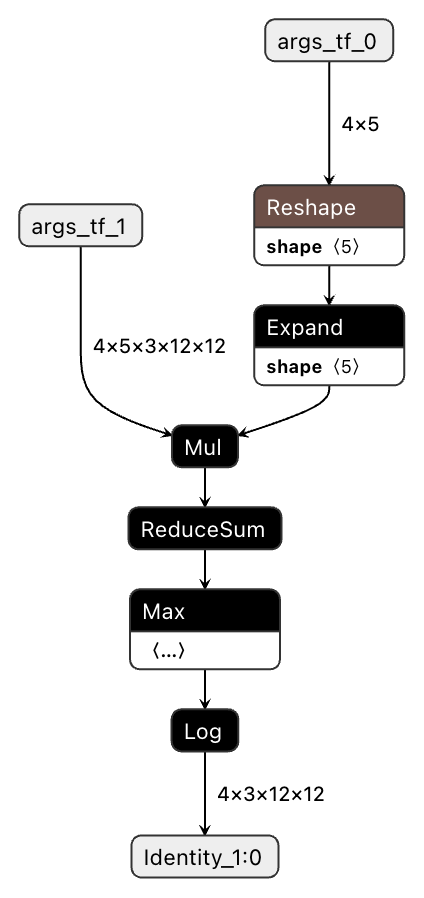}
        \caption{Our unified sparse graph.}
        \label{fig:right}
    \end{subfigure}
    \vspace{-0.2cm}
    \caption{Our methodology generates structured computational graphs, enabling efficient mapping to GPU kernels and facilitating HW acceleration.}
    \label{fig:structured_graph}
\end{figure}

\noindent\textbf{Proposed methodology}. We remodel \textit{pymdp} to generate a unified, sparse structure, which leaves all probabilistic computations mathematically unchanged:
\vspace{-0.6cm}
\begin{enumerate}\setlength{\itemsep}{2pt}
    \item \textbf{Unified dense view.}  
    All factors are packed into shape-aligned, padded arrays, allowing %state/policy 
    inference routines to be expressed as broadcasted tensor operations—removing \textit{for}-loops and enabling efficient vectorization, see Fig~\ref{fig:structured_graph}.
    \item \textbf{Restoring sparsity.}  
    We then replace dense arrays with \textit{JAX BCOO} objects \cite{jax2018github},  %This captures 
    capturing both \emph{structural sparsity} (missing links) and \emph{functional sparsity} while preserving the unified computational graph obtained in step 1.
\end{enumerate}

\vspace{-0.7cm}
\section{Results and Discussion}
\vspace{-0.3cm}
%\begin{itemize}
%    \item Mention the environments or tasks used for testing (e.g., simulated agents, grid worlds, robotics).
%    \item Present key findings (e.g., performance metrics, trade-offs).
%    \item Discuss implications for real-time or embedded applications. Highlight any surprising results or limitations.
%    \item Suggest directions for future research (e.g., GPU acceleration, cross-language integration, real-world deployment).
%\end{itemize}

\begin{figure}[t!]
    \centering
    \begin{subfigure}[t]{0.49\textwidth}
        \centering
        \includegraphics[width=\linewidth, height=4cm]{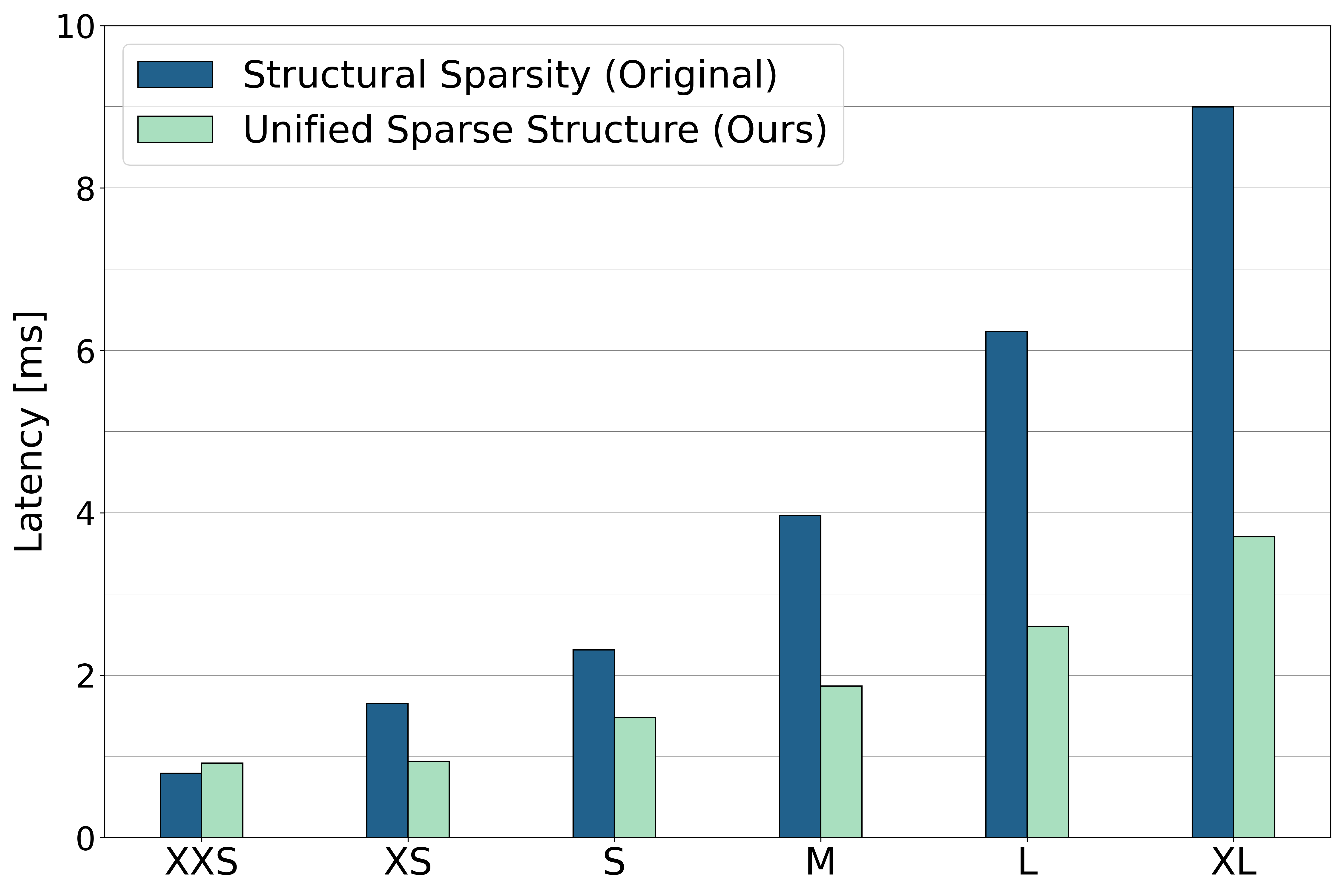}
        \caption{Latency comparison (ms).}
        \label{fig:latency}
    \end{subfigure}
    \hfill
    \begin{subfigure}[t]{0.49\textwidth}
        \centering
        \includegraphics[width=\linewidth, height=4cm]{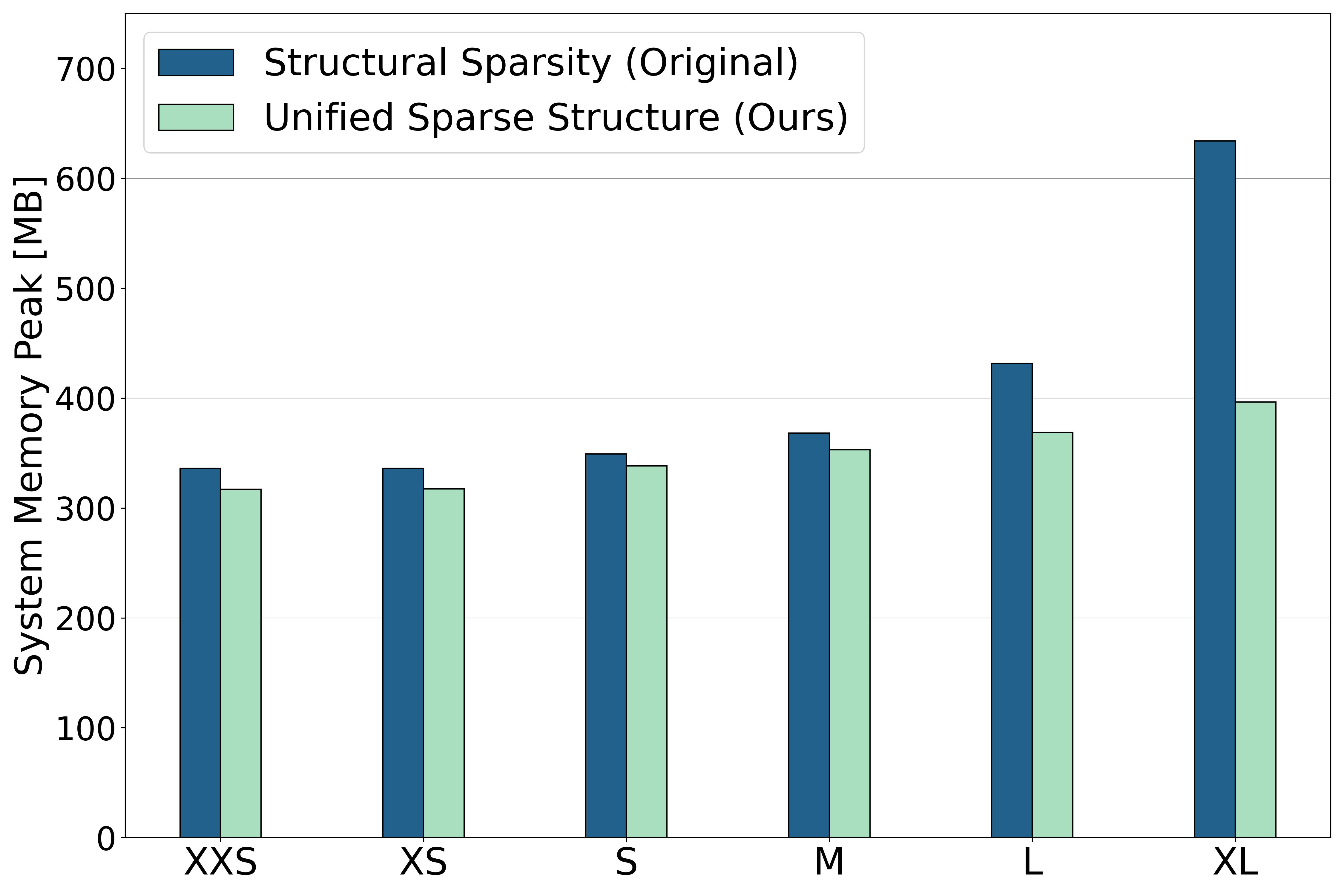}
        \caption{System memory comparison (MB)}
        \label{fig:memory}
    \end{subfigure}
    %\vspace{-0.2cm}
    \caption{Comparison of the \textit{log-likelihood computation} when our methodology is applied on a 100-run benchmark with a warm-up sample on an Nvidia Jetson Orin AGX, a high-end embedded device, featuring a multi-core CPU and an embedded GPU for robotics and edge AI applications. We parametrize a set of generative models sizing from \textit{XXS} to \textit{XL}, whose parameters are in Table \ref{model-parameters}.}
    \label{fig:benchmark}
\end{figure}

We apply the proposed methodology to a core computation used by \texttt{pymdp}'s inference routines, the \textit{log-likelihood} method, demonstrating the practical effectiveness of our ongoing work on a set of parametrized AIF agents (Table~\ref{model-parameters}). 
%All experiments have been performed on an Nvidia Jetson Orin AGX, a high-end embedded device, featuring a multi-core CPU and an embedded GPU for robotics and edge AI applications. % on a 100-run benchmark with a warm-up sample.
Figure \ref{fig:latency} compares the \textit{log-likelihood computation} latency between the current implementations in \textit{pymdp} and our proposed approach. Our unified implementation notably outperforms the baseline thanks to its compressed representation and efficient HW mapping, scaling significantly better and achieving speed-ups of over 2x. Even though our approach requires specifying model parameters in a way that incurs a higher parameter count, we are able to exploit sparsity to a larger degree, leading to fewer effective parameters. This is demonstrated in figure~\ref{fig:memory}, where a reduction of up to 35\% in system memory is accomplished.% during model execution.

Overall, our methodology establishes a path for deployment in \textit{edge} devices by uniting \textit{pymdp}'s flexibility, \textit{JAX}'s efficiency, and optimized computational graphs for HW acceleration. We are actively extending support to the full \textit{pymdp} API and envision deployment on ultra-low-power platforms. %by exporting the graph to intermediate representations, e.g., ONNX or MLIR. 

\vspace{-1cm}

\begin{table}[t!]
    % \vspace{-0.2cm}
    \centering
    \small
    \setlength{\tabcolsep}{3pt}
    \begin{tabular}{
        >{\centering\arraybackslash}p{0.10\linewidth}  % Model
        >{\centering\arraybackslash}p{0.15\linewidth}  % Type (Original/Sparse)
        >{\centering\arraybackslash}p{0.12\linewidth}  % Modalities
        %>{\centering\arraybackslash}p{0.13\linewidth}  % A Deps Length
        >{\centering\arraybackslash}p{0.13\linewidth}  % Hidden States
        >{\centering\arraybackslash}p{0.2 \linewidth}  % Params
        >{\centering\arraybackslash}p{0.2\linewidth}  % Sparsity
    }
        \makecell{\textbf{Model}} & \makecell{\textbf{Type}} & \makecell{\textbf{\#Moda-}\\\textbf{lities}} &
        %\makecell{\textbf{A Deps}\\\textbf{Length}} &
        \makecell{\textbf{\#Hidden}\\\textbf{States}} &
        \makecell{\textbf{\#Params}\\\textbf{(M)}} &
        \makecell{\textbf{Sparsity}\\\textbf{(\%)}} \\
        \hline
        XXS & Orig. (Ours) & 16 & 60 & 0.003 (0.010) & 61.0 (89.4)\\
        XS & Orig. (Ours) & 46 & 180 & 0.026 (0.279) & 59.7 (96.2)\\
        S & Orig. (Ours)& 92 &  364 & 0.108 (2.285) & 57.6 (98.0) \\
        M & Orig. (Ours) & 154 & 612 & 0.304 (10.81) & 56.4 (98.7) \\
        L & Orig. (Ours) & 232 & 924 & 0.694 (37.13) & 55.7 (99.2)  \\
        XL & Orig. (Ours) & 326 & 1300 & 1.373 (103.3) & 55.2 (99.4)\\
    \end{tabular}
    \vspace{0.2cm}
    \caption[Models' parameters for log-likelihood computation]{
    Models' parameters associated with the log-likelihood computation. %\textit{A Dependency Length} indicates the relationship between factors and modalities\footnotemark[1].
    }
    \label{model-parameters}
\end{table}
%\footnotetext[1]{\emph{A Dependency Length} is defined as $\frac{1}{\#\text{Modalities}} \sum_{i = 0}^{\#\text{Modalities} - 1} \text{len}(A\_\text{dependencies}[i])$.}

% 1.937 & 1.978 & 1.989 & 1.993 & 1.995 & 1.997 &
\newpage

\begin{credits}
\subsubsection{\ackname} This work was partly supported by Horizon Europe dAIEdge under grant No. 101120726.
\end{credits}
\bibliographystyle{splncs04}
\bibliography{bibs}

\end{document}